\documentclass[lettersize,journal]{IEEEtran}
\usepackage{amsmath,amsfonts}
\usepackage{algorithmic}
\usepackage{algorithm}
\usepackage{array}

\usepackage[caption=false,font=normalsize,labelfont=sf,textfont=sf]{subfig}
\usepackage{textcomp}
\usepackage{stfloats}
\usepackage{url}
\usepackage{verbatim}
\usepackage{graphicx}
\usepackage{cite}
\usepackage{booktabs}
\usepackage{bm}
\usepackage{multirow}

\usepackage{tikz}
\usepackage{comment}
\usepackage{amsmath,amssymb} 
\usepackage{color}
\definecolor{dodgeblue}{RGB}{30,144,255}
\definecolor{lightgreen}{RGB}{0,157,0}

\usepackage{diagbox}

\usepackage[breaklinks=true,colorlinks,bookmarks=false]{hyperref}

\usepackage{colortbl}
\definecolor{mygray}{gray}{.9}
\usepackage{array, boldline, makecell, booktabs}

\usepackage{xcolor}
\definecolor{firered}{RGB}{222,82,57}
\definecolor{iceblue}{RGB}{33,102,200}
\usepackage{multirow}
\usepackage{makecell}
\usepackage{diagbox} 
\definecolor{mygray}{gray}{.9}
\usepackage{adjustbox}

\usepackage{textcomp}
\usepackage{stfloats}
\usepackage{bm}
\makeatletter
\newcommand{\thickhline}{%
    \noalign {\ifnum 0=`}\fi \hrule height 1pt
    \futurelet \reserved@a \@xhline
}

\begin{document}

\title{CML-MOTS: Collaborative Multi-task Learning for Multi-Object Tracking and Segmentation}

\author{Yiming Cui, Cheng Han, Dongfang Liu 

\IEEEcompsocitemizethanks{
\IEEEcompsocthanksitem Yiming Cui is with the University of Florida, Gainesville, FL 32611, USA.
\IEEEcompsocthanksitem Cheng Han and Dongfang Liu are with the Rochester Institute of Technology, Rochester, New York, USA.}
}
\maketitle
\vspace{-2.6cm}
\begin{abstract}
 The advancement of computer vision has pushed visual analysis tasks from still images to the video domain. In recent years, video instance segmentation, which aims to track and segment multiple objects in video frames, has drawn much attention for its potential applications in various emerging areas such as autonomous driving, intelligent transportation, and smart retail. In this paper, we propose an effective framework for instance-level visual analysis on video frames, which can simultaneously conduct object detection, instance segmentation, and multi-object tracking. The core idea of our method is collaborative multi-task learning which is achieved by a novel structure, named associative connections among detection, segmentation, and tracking task heads in an end-to-end learnable CNN. These additional connections allow information propagation across multiple related tasks, to benefit these tasks simultaneously. We evaluate the proposed method extensively on KITTI MOTS and MOTS Challenge datasets and obtain quite encouraging results.
\end{abstract}
%
%
%
\vspace{-0.8cm}
\section{Introduction}
In the past decade, the computer vision community has achieved significant progress in many tasks with the development of deep learning~\cite{LIU201941,7959631,han2015csvt,huang19tsmc}. Among various visual tasks, instance segmentation~\cite{he2017mask} has drawn wide attention due to its importance in many emerging applications, such as autonomous driving \cite{liu2021visual, liu2022prophet, yan2021hierarchical, liu2021densernet, liu2020indoor, wang2019end}, augmented reality \cite{cao2021vector,  cao2022towards}, and video captioning \cite{Yan2022GLRGGR, yan2022video}. Technically, it is quite challenging as it is a compound task consisting of both object detection and segmentation, each of which is a difficult task and has been studied for a long time. 
	
\textcolor{black}{Compared to instance segmentation on images, multi-object tracking, and segmentation is much more challenging because it not only needs to perform instance-level segmentation on individual frames but also has to depict the coherence of each instance in consecutive video frames~\cite{bertasius2020classifying}. Due to these challenges, multi-object tracking and segmentation have received much attention in recent years~\cite{yang2019video,cui2022dg, voigtlaender2019mots, liu2021sg, yang2020remots, xu2020pointtrack++}. In general, multi-object tracking and segmentation contain object detection, segmentation, and tracking simultaneously in consecutive video frames. Compared to video object segmentation~\cite{wang2019fast} that deals with object segmentation and tracking in videos, multi-object tracking and segmentation require additional object detection. It also has to generate object masks compared to video object detection which contains both object detection and tracking in videos~\cite{zhu2017deep,cui2022dynamic, wang2022towards, cui2022dfa, zhu2017flow, cui2021tf}.} 

Currently, the state-of-the-art methods are mainly based on Mask R-CNN~\cite{he2017mask} while adding tracking sub-network. The Mask R-CNN family~\cite{xu2020pointtrack++,voigtlaender2019mots} has demonstrated an appealing performance on this task. However, these methods still have several drawbacks. For instance, TrackR-CNN~\cite{voigtlaender2019mots} uses proposal-based ROI features, which are not fine enough for mask and tracking heads to produce accurate predictions. \textcolor{black}{In detail, the current methods use the coarse proposal-based ROI features directly, which is not enough. Instead, our method processes the coarse proposal-based ROI features first and then uses the refined version for the downstream tasks.} In addition, using proposal-based features not only has high computational complexity but also makes it prone to incorrect or redundant predictions. What is more, TrackR-CNN models object movements and scene consistency among video frames by using a 3D convolutional operation, which is also parameter-heavy and computationally expensive. Although PointTrack~\cite{xu2020pointtrack++} achieves a significant improvement in segmentation results by proposing a more powerful mask head, it is a \textcolor{black}{multi-step} architecture and so cannot be jointly optimized. \textcolor{black}{In PointTrack, detection, and segmentation operations are applied to the input first. This partial model is optimized first without the tracking task. Then the detected and segmented objects are transformed into point cloud formats for the tracking task, which is optimized. In general, the whole model contains multiple \textcolor{black}{steps} and is not optimized jointly or end-to-end trainable.} Moreover, its point-cloud strategy requires additional post-processing. Meanwhile, although these methods are trained by multiple learning objectives corresponding to different tasks, the relations among these tasks have not been explored.	

Intuitively, object detection could benefit instance segmentation, and good object masks are also helpful for multi-object tracking. Inspired by this intuition, we propose a novel idea of collaborative multi-task learning for multi-object tracking and segmentation. Associative connections are added among different task heads to enable information propagation across them. Although our method is also based on Mask R-CNN~\cite{he2017mask}, it fundamentally differs from other Mask R-CNN variants~\cite{voigtlaender2019mots,xu2020pointtrack++} as our method exploits associative connections to facilitate accurate information propagation across the detection, segmentation, and tracking heads to benefit individual tasks. With associative connections, our segmentation and tracking can perform predictions on refined ROI features instead of using features pooled from the very coarse region proposals. Extensive experimental results demonstrate significant improvements over the existing state-of-the-art methods for multi-object tracking and segmentation on two benchmarks~(KITTI MOTS~\cite{voigtlaender2019mots} and MOTS Challenge~\cite{milan2016mot16}). The principal contributions of this work can be summarized as follows,
\begin{itemize}
    \item We propose a novel idea of collaborative multi-task learning for multiple object tracking and segmentation in videos.
    \item We design associative connections among different tasks to enable information flow through different task heads, to simultaneously learn from multiple tasks and consider their intrinsic relations in the meantime.
    \item Compared to existing methods, our method achieves better results on KITTI MOTS and MOTS Challenge benchmarks, especially on the consistency of multi-object tracking.
\end{itemize}
\vspace{-0.4cm}
\section{Related Work}
Visual tasks in the video domain have been under-explored in the literature compared to image-level tasks. Particularly, video instance segmentation which includes detection, segmentation, and tracking is largely ignored due to its extreme difficulties. This section offers a review of recent works about several related tasks to video instance segmentation.
\begin{figure*}[!ht]
		\centering
    \includegraphics[width=\textwidth]{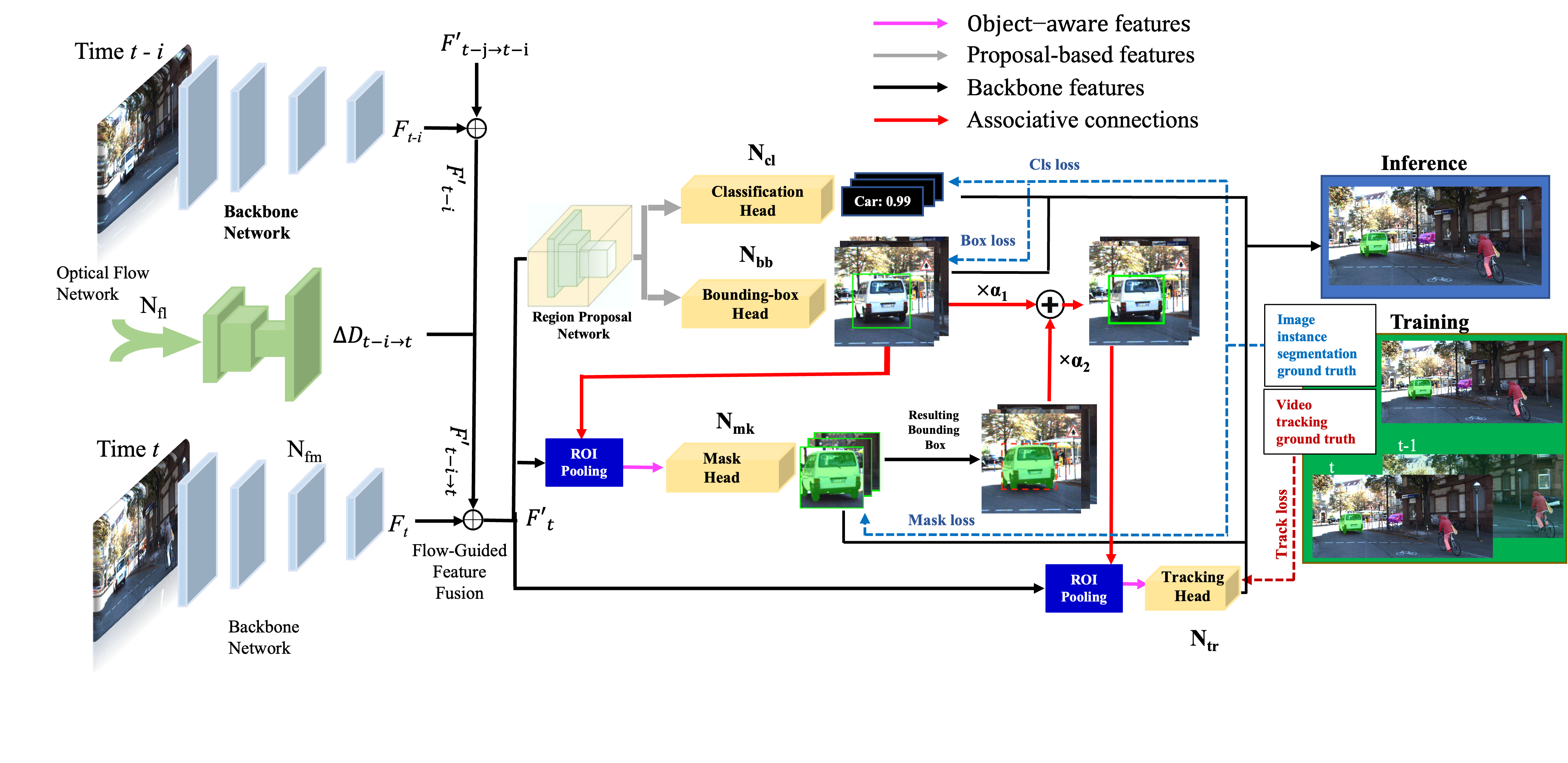}
    \vspace{-1.8cm}
    \caption{Illustration of the proposed framework. It adds associative connections among different task heads in TrackR-CNN~\cite{voigtlaender2019mots} to enable collaboration among multiple tasks by facilitating information propagation across them. Particularly,  the bounding boxes outputted by the detection head are linked to the mask head to extract object-aware features for instance segmentation. The outputs of the mask head are fused with the outputs of the detection head to obtain more reliable bounding boxes, which are then inputted into the tracking head for better tracking ability. Besides, a lightweight optical flow network is used to model the object movements across successive frames, which is used for aligning object features among these frames to obtain more powerful feature maps in the currently processed frame.}
    \label{Frame}
\end{figure*}
\vspace{-0.4cm}
\subsection{Image Object Detection}
State-of-the-art object detection methods~\cite{ren2015faster, object_detection_tnnls19, cui2022dynamic2, liu2020large, yang2023techniques, zhang2018improved, cui2022dynamic, dong2023watchdog} are generally based on deep CNNs for feature extraction and a shallow detection structure for detection prediction, including classification and bounding box regression. R-CNN~\cite{girshick2014rich} proposed a multi-stage pipeline to classify region proposals at different semantic levels for object detection. To speed up, Fast R-CNN~\cite{girshick2015fast} and Libra R-CNN~\cite{pang2019libra} used ROI pooling on the feature maps which are shared on the entire image. As a representative work of the multi-stage detection family, Faster R-CNN~\cite{ren2015faster} introduced a Region Proposal Network (RPN) to generate region proposals and then the proposal-based features are shared between classification and bounding box regression heads. R-FCN~\cite{dai2016r} replaced ROI pooling with position-sensitivity ROI pooling to further improve the recognition accuracy while still facilitating feature sharing.

\textcolor{black}{Traditional two-stage object detectors, exemplified by the R-CNN family \cite{Ren_2017,girshick2015fast,ren2015faster}, rely on a plethora of predefined anchor boxes to designate initial object candidate locations. In contrast, one-stage methods \cite{redmon2016you,bochkovskiy2020yolov4} were introduced to enhance the efficiency and inference speed of object detectors by forgoing the use of region proposals. Recently, query-based approaches \cite{detr,zhu2021deformable, cui2021geometric,sun2021sparse} have emerged, replacing anchor boxes and region proposals with learned proposals or queries. DETR \cite{detr} adapts an encoder-decoder architecture based on transformers \cite{vaswani2017attention} to generate a sequence of prediction outputs. It introduces a set loss function to facilitate bipartite matching between predicted and ground-truth objects. Deformable-DETR \cite{zhu2021deformable} enhances the convergence of DETR by refining feature spatial resolutions. Sparse R-CNN \cite{sun2021sparse} employs a fixed sparse set of learned object proposals to classify and localize objects in the image. It utilizes dynamic heads to generate final predictions directly, eliminating the need for post-processing techniques like non-maximum suppression.}
\vspace{-0.4cm}
\subsection{Image Instance Segmentation}
Instance segmentation not only predicts semantic classes on each pixel but also groups pixels into different object instances. Due to the effectiveness of R-CNN\cite{girshick2014rich}, many instance segmentation methods perform mask prediction on top of region proposals. Some early methods are based on the bottom-up segment strategy. For instance, the DeepMask family~\cite{yang2019video,chen2019hybrid} learns to segment proposal candidates first and then classify them using Fast R-CNN. These methods have segmentation precede detection, which is slow and inaccurate. 
	
Another strategy for instance segmentation is based on a parallel prediction of masks and class labels~\cite{he2017mask}. Li et al.~\cite{li2017fully} proposed the fully convolutional instance segmentation by combining the segmentation proposal system in~\cite{chen2019hybrid} and object detection system in R-FCN~\cite{dai2016r}. Basically, \cite{chen2019hybrid,li2017fully} all use a set of position-sensitive output channels to simultaneously predict object classes, bounding boxes, and masks, thus being fast and fundamentally similar. However, they struggle to deal with occlusion or truncation instances as they tend to create spurious edges in those cases. 
	
\textcolor{black}{Besides two-stage methods, one-stage instance segmentation frameworks like YOLACT \cite{yolact-iccv2019,yolact-plus-tpami2020}, SipMask \cite{Cao_SipMask_ECCV_2020}, and SOLO \cite{wang2020solo,wang2020solov2} have been introduced to strike a balance between inference speed and accuracy. Recently, QueryInst \cite{Fang_2021_ICCV} extended the query-based object detection method Sparse R-CNN \cite{sun2021sparse} to the instance segmentation task by incorporating a dynamic mask head and parallel supervision. Nevertheless, all the two-stage and query-based methods mentioned earlier employ a fixed number of proposals, which may not be adaptable to images with varying objects or devices with different computational resource constraints.}
\vspace{-0.4cm}
\subsection{Object Tracking}
Tracking-by-detection is a popular strategy for multi-object tracking~\cite{liu15tnnls,wang2017tip,chu2019famnet}. A common practice is to associate the tracks with detection based on their confidences~\cite{chu2019famnet}. To improve the tracking accuracy, Sun et al.~\cite{sun2019deep} exploited multiple detectors by considering outputs from multiple over-detected detectors. However, this method enhances tracking performance at the cost of high computational complexity. More recently, Kim et al.~\cite{kim2018multi} proposed to use a single object tracker based on a binary classifier for online multi-object tracking. Their proposed architecture has shared features for tracking and classification to speed up the process. Even though, \cite{kim2018multi} is still computationally expensive for real-time applications. 
	
Many previous methods deal with tracking tasks as a global optimization problem \cite{yang2019video,tang2017multiple}. This kind of method formulates temporal information from nearby frames to reduce noisy detection and handle ambiguities for object association. The principal strategy is to re-identify objects using the embedding vectors since each association vector represents the identity of an object~\cite{voigtlaender2019mots,tang2017multiple}. Inspired by the aforementioned methods, our work also leverages deeply learned ReID features. With the proposed associative connections, our work has better object-aware features to obtain improved identification performance. 
\vspace{-0.8cm}
\subsection{Video Object Detection}
\textcolor{black}{Video object detection involves the identification and localization of objects of interest in each frame, even in the presence of potential degradation due to rapid motion. Present approaches \cite{kang2016object, han2016seq, kang2017t, wu2019sequence, cui2023dq, zhu2017flow, zhu2018towards, chen2020memory, wang2018fully, cui2021tf, cui2022dfa, cui2023feature} typically extend image-based object detectors into the realm of videos. These models fall into two categories: Post-processing-based and feature-aggregation-based.}

\textcolor{black}{Post-processing-based models extend image object detectors to video by linking prediction results across frames based on temporal relationships \cite{kang2016object, han2016seq, kang2017t}. Examples include T-CNN \cite{kang2016object} and Seq-NMS \cite{han2016seq}. T-CNN employs a CNN-based pipeline with straightforward object tracking for video object detection. Seq-NMS associates prediction results from each frame using the IOU threshold. While these models outperform single-image object detectors, they heavily rely on individual frame detections and lack joint optimization. If single-frame outputs are erroneous, the post-processing pipeline cannot rectify them, resulting in suboptimal performance. Furthermore, these models tend to be slower as they process each frame independently before post-processing.}

\textcolor{black}{In contrast, feature-aggregation-based and Transformer-based models can aggregate information across frames and jointly optimize predictions, yielding improved performance. These models effectively utilize temporal and spatial cues to track and detect objects across consecutive frames, making them better suited for video object detection tasks. While post-processing-based models offer some improvement over single-image object detectors, they are constrained by their reliance on individual frame results and slower inference. Feature-aggregation-based and Transformer-based models present a more promising approach, leveraging data from multiple frames and optimizing jointly.}

\textcolor{black}{Feature-aggregation-based models enhance current frame representations by incorporating features from adjacent frames, assuming they can mitigate feature degradation. Several models embody this concept. For instance, FGFA \cite{zhu2017flow} employs estimated optical flow to fuse neighboring features, while MANet combines pixel-level and instance-level object features. Conversely, SELSA \cite{wu2019sequence} calibrates features based on semantic similarity rather than temporal relations. MEGA \cite{chen2020memory} integrates local and global temporal information to enhance performance. Although these models surpass post-processing-based ones in performance, they typically demand more computational resources, resulting in slower inference speeds. In summary, while feature-aggregation-based models have enhanced video object detection, they often come at the expense of slower inference. Transformer-based models offer a promising solution to this issue by efficiently fusing information across frames, suggesting further advancements in video object detection tasks.}
\vspace{-0.4cm}
\subsection{Video Instance Segmentation}
\textcolor{black}{Recent endeavors on video instance segmentation  \cite{voigtlaender2019mots,xu2020pointtrack++,yang2019video} are intuitive extensions of Mask R-CNN~\cite{he2017mask} while considering additional motion cues~\cite{cheng2017segflow} or temporal consistency \cite{perazzi2017learning, elallid2022deep, yang2018efficient} in videos.	}
\textcolor{black}{There are three main categories of existing methods for Visual Instance Segmentation (VIS): two-stage, one-stage, and transformer-based. Two-stage approaches \cite{bertasius2020classifying} build upon the Mask R-CNN family \cite{he2017mask, ren2015faster}, incorporating an additional tracking branch for object association. One-stage methods \cite{liu2021sg, cao2020sipmask} employ anchor-free detectors \cite{tian2019fcos}, often utilizing linear mask basis combination \cite{yolact-iccv2019} or conditional mask prediction generation \cite{tian2020conditional}. Transformer-based models \cite{cheng2021mask2former, heo2022vita, wang2021end, yang2022temporally} introduce innovative adaptations of the transformer architecture for VIS tasks. VisTr \cite{wang2021end} pioneers the application of transformers in VIS, and IFC \cite{hwang2021video} enhances efficiency through the use of memory tokens. Seqformer \cite{wu2022seqformer} introduces frame query decomposition, while Mask2Former \cite{cheng2021mask2former} incorporates masked attention. VMT \cite{ke2022video} extends the Mask Transfiner \cite{ke2022mask} to video for high-quality VIS, and IDOL \cite{wu2022defense} specializes in online VIS.}

\subsection{Multi-object Tracking and Segmentation}
\textcolor{black}{Multi-object tracking (MOT) is a crucial task in autonomous driving, encompassing both object detection and tracking within video sequences. Numerous datasets have been curated with a focus on driving scenarios, including KITTI tracking~\cite{geiger2012we}, MOTChallenge~\cite{milan2016mot16}, UA-DETRAC~\cite{wen2015ua}, PathTrack~\cite{manen2017pathtrack}, and PoseTrack~\cite{andriluka2018posetrack}. However, none of these datasets offer segmentation masks for annotated objects, thus lacking pixel-level representations and intricate interactions seen in MOTS data. More advanced datasets, such as Cityscapes~\cite{cordts2015cityscapes}, ApolloScape~\cite{huang2018apolloscape}, BDD100K~\cite{yu2020bdd100k}, and KITTI MOTS dataset~\cite{voigtlaender2019mots}, do provide instance segmentation data for autonomous driving. Nevertheless, Cityscapes only supplies instance annotations for a small subset (i.e., 5,000 images), and ApolloScape does not offer temporal object descriptions over time. Consequently, neither dataset is suitable for the joint training of MOTS algorithms. In contrast, KITTI MOTS~\cite{voigtlaender2019mots} stands as the first public dataset that addresses the scarcity of data for the MOTS task, albeit with a relatively modest number of training samples. To date, BDD100K boasts the largest scale of data from intensive sequential frames, which may be considered redundant for training purposes. In comparison to the aforementioned datasets, our DGL-MOTS dataset encompasses a wider range of diverse data with finely detailed annotations.}

\vspace{-0.4cm}
\section{The Proposed Method}
\subsection{Overview}
Fig.~\ref{Frame} shows the proposed network for multi-objects tracking and segmentation. It follows the basic network architectures for object detection, instance segmentation, and object tracking, containing two parts: one for extracting feature maps and the other for different tasks sharing the extracted features as inputs. More specifically, the first part of our method consists of two components: a backbone feature extraction network $\textbf{N}_{fm}$ to compute per frame feature maps and an optical flow network $\textbf{N}_{fl}$ to estimate object movements across nearby frames. With the help of optical flow, feature maps from previous frames can be warped into the current frame, which is used to combine with the extracted feature maps at the current frame, to obtain an enhanced feature representation of the current frame that should be more robust to image blur, occlusion, etc. Accordingly, the second part of our method is constituted of three heads, each of which corresponds to a specific task, i.e., object detection, instance segmentation, and multi-object tracking. The detection head contains classification head $\textbf{N}_{cl}$ and bounding-box regression head $\textbf{N}_{bb}$ while the instance segmentation is achieved by the mask prediction head $\textbf{N}_{mk}$. The tracking head $\textbf{N}_{tr}$ aims to identify the same objects that appeared in multiple frames. 
	
To train such a network with different heads end-to-end, existing methods resort to multi-task learning that simultaneously optimizes losses related to individual tasks. However, those methods ignore the intrinsic correlations among these tasks, which could benefit each other if used properly. \textcolor{black}{In detail, each loss is only designed and optimized specifically for one task and there is no interaction between different tasks and their corresponding losses. In other words, the performance of the instance segmentation task will not affect that of the object detection task and vice versa. In this case, we argue that the designs of losses can be optimized.} Therefore, we propose collaborative multi-task learning to enable information propagation among these individual tasks when optimizing the total learning objective containing all these tasks. For this purpose, we introduce associative connections among detection, segmentation, and tracking heads so that the network training could be aware of the interactions among these three tasks. The introduced associative connections are shown by the red arrows in Fig.~\ref{Frame}. Besides enabling information propagation among different task heads, our network is more efficient compared to previous methods. Existing MOTS methods~\cite{voigtlaender2019mots,luiten2020unovost,lin2019agss} that also use three task heads encounter the problem of computing redundant feature representations, as they rely on proposal features for instance segmentation and tracking. With the help of associative connections, our segmentation and tracking heads only take the detected bounding boxes or masks for input, thus avoiding computing features on unrelated proposals. In addition, due to the added associative connections, the mask head and tracking head can use more accurate features extracted on object bounding boxes instead of the region proposals. Therefore, higher-quality instance segmentation and tracking results can be expected.
	
In the following subsections, we describe in detail the proposed method, including feature extraction network, associative connections, training, and inference, as well as the network architecture. Table~\ref{tab:notation} lists the main symbols used in this paper for the neatness of description.

\begin{table*}[!ht]
\vspace{-0.3cm}
    \caption{Summarization of the main notations used in this paper.}
    \label{tab:notation}
    \centering
\vspace{-0.3cm}    
    \begin{tabular}{p{1.2cm}<{\centering}|p{9.8cm}<{\centering}}
        \toprule
        \multicolumn{2}{c}{Notation} \\
        \midrule
        $\textbf{N}_{fm}$ & Feature extraction network \\
        $\textbf{N}_{fl}$ & Optical flow network \\
        $\textbf{N}_{cl}$ & Classification head \\
        $\textbf{N}_{bb}$ & Bounding box head \\
        $\textbf{N}_{mk}$ & Mask  head \\
        $\textbf{N}_{tr}$ & Tracking head \\
        \midrule
        $F_t$ & Feature maps at time $t$ \\
        $F'_t$ & Enhanced feature maps at time $t$ \\
        $I_t$ & Input frame at time $t$ \\
        ${F}_{t-i \rightarrow t}$ & Warped feature maps from time $t-i$ to $t$ \\
        $\Delta D_{t-i \rightarrow t}$ & Object movement between $t$ and $t-i$\\
        ${\omega}_{k \rightarrow t}$ & Weight for fusing feature of the $k$th frame at the time $t$\\
        \midrule
        ${b}^{i}$ & The bounding box of the ${i}^{th}$ object \\
        ${b}_{mk}^{i}$ & The bounding box computed from the predicted mask of the $i^{th}$ object \\
        ${b}_{wb}^{i}$ & Weighted bounding box of the ${i}^{th}$ object connecting the detection and segmentation heads to the tracking head \\
        \midrule
        $\mathcal{L}_{cls}$ & Loss for classification \\
        $\mathcal{L}_{box}$ & Loss for bounding box regression \\
        $\mathcal{L}_{mask}$ & Loss for mask generation \\
        $\mathcal{L}_{track}$ & Loss for tracking \\
        $\mathcal{L}_{total}$ & Total loss for training \\
        \bottomrule
    \end{tabular}
\end{table*}
\vspace{-0.4cm}
\subsection{Network Architecture}
Our network architectures have general and flexible design options. We craft state-of-the-art architectures into the proposed methods for different visual tasks. Particularly, the proposed method has three contributing modules: (i) the CNN backbone architecture employed for feature extraction over the input frame, (ii) the optical flow architecture used for motion estimation across frames, and (iii) the task heads for classification, location regression (bounding box), mask generation, and object tracking. With associative connections, the mask head and tracking head are applied to object-aware RoI instead of using RoI from proposals.

\subsubsection{Feature extraction}	
 The ResNet-101~\cite{he2016deep} is used as our backbone feature extraction network $\textbf{N}_{fm}$ in this paper. According to the practice of using ResNet-101 as the backbone in Faster R-CNN, the outputs of its final convolutional layer $C4$ are feedforwarded to the task heads.

\subsubsection{Motion estimation}
The simple version of FlowNet~\cite{dosovitskiy2015flownet} is used as the flow network $\textbf{N}_{fl}$ for motion estimation across video frames. It is pretrained on the synthetic Flying Chairs dataset~\cite{dosovitskiy2015flownet}.
According to \cite{zhu2017deep,zhu2017flow}, we have the input frame half-sized and the output stride 4. Therefore, the output resolution of the generated flow field is $1/8$ of the original frame size. 
Since the output of the feature extraction network has a stride of 16, we use bilinear interpolation to further down-sample the flow field and scale the field by half, to match the resolution of extracted feature maps. The down-sample process is non-learnable as the bilinear interpolation is a parameter-free layer in the network and is also differentiated during training.
	
\subsubsection{Task heads}
For the task heads, we follow architectures presented in Faster R-CNN~\cite{ren2015faster}, Mask RCNN~\cite{he2017mask}, and TrackR-CNN~\cite{voigtlaender2019mots} for different tasks. For detection, we craft Faster R-CNN, a two-stage detector for object classification and bounding box regression. For mask prediction, we follow  Mask RCNN by adding a fully convolutional mask prediction branch. For the tracking task, we include an association layer~\cite{voigtlaender2019mots} to calculate the distance of 128-D identity vectors to track different objects across frames. Among these heads, there are associative connections to collect object-aware features and propagate them across these tasks. Introducing associative connections among other task heads to facilitate information flow interactively among several tasks is also the main contribution of this work. Compared to previous proposal-based features, this work uses object-aware features by leveraging associative connections, which shows better performance in predicting masks and tracking the same identities.
 \vspace{-0.4cm}
\subsection{Feature Extraction Guided by Optical Flow}\label{sec:feat}
Given an input frame $I_t$ at time $t$, the feature extraction process can be expressed as $F_t=\textbf{N}_{fm}(I_t)$. Note that $F_t$ are only intermediate feature maps, which will be enhanced with features from previous frames based on the estimated object movements by $\textbf{N}_{fl}$. It is the enhanced feature maps being passed into the following heads regarding different tasks. Although various backbone networks can be used here for feature extraction, we use ResNet-101~\cite{he2016deep} in this paper due to its popularity.
	
To enhance the feature representation of different objects on the input frame by leveraging on its previous frames, we exploit the temporal visual cues with the help of an optical flow network $\textbf{N}_{fl}$. Thus, the enhanced feature representation is also called flow-guided features. The warped feature from time $t-i$ to $t$ is denote as: 
\begin{equation}
\begin{aligned}
F_{t-i \rightarrow t}=\mathcal{WP}(F_{t-i},\Delta D_{t-i \rightarrow t}),
\end{aligned}
\label{Equation 1}
\end{equation}
where $\mathcal{WP}$ is the feature warping function to predict the feature maps at $t$ based on the feature maps at $t-i$~(i.e., $F_{t-i}$) and the estimated object movement $\Delta D_{t-i \rightarrow t}$ between $t$ and $t-i$ frames. The movement is predicted by the optical flow network, i.e., $\Delta D_{t-i \rightarrow t} = \textbf{N}_{fl}(I_{t-i},I_{t})$. By modeling the feature map movements from nearby frames, we hope to improve the extracted features that may have been originally compromised by motion blur, defocus, or occlusion that often happened in the video domain.
	
Since in real scenarios processing a specific video frame can only rely on its previous frames, given an input frame $I_t$, we obtain a set of warped feature maps for each of its previous frames to compute the enhanced feature maps of the current frame $I_t$. Specifically, with a predefined temporal range $n$, each feature map of the previous frames in this range is warped into frame $t$ according to Eq.~(\ref{Equation 1}), resulting in a set of predicted feature maps $\{F_{t-i\rightarrow t}|i\in [1, n]\}$. Note that we warp previous features to every current frame for feature fusion instead of keyframes or using dense aggregation~\cite{zhu2017flow,zhu2017deep}. Due to the high efficiency of the used flow network~(i.e., FlowNet~\cite{dosovitskiy2015flownet}), this enables our method to extract strong features but still with affordable computational cost. Runtime analysis is provided in Section~\ref{sec:time}.

Given the set of warped feature maps $\{F_{t-i\rightarrow t}|i\in [1, n]\}$, the fused feature  maps $F'_{t}$ at the frame $I_t$ is then computed by the weighted average of these warped features and $F_{t}$, 
\begin{equation}
\begin{aligned}
F'_{t}=\textstyle\sum\limits_{k \in [t-n,t-1]}\big(\omega_{{k}\rightarrow t}\cdot F'_{{k}\rightarrow t}\big)+F_{t},
\end{aligned}
\label{Equation 2}
\end{equation}
where the weight $\omega_{{k}\rightarrow t}$ is adaptively computed based on the similarity between $F_{k\rightarrow t}$ and $F_{t}$. Among many choices for defining the similarity between these feature maps, we follow the implementation in~\cite{liu2020video} to use a shallow fully convolutional network to output embedding vectors for similarity computation. That is, $\omega_{{k}\rightarrow t}$ is computed as,
\begin{equation}
\omega_{k\rightarrow t} = \text{exp}\bigg(
\frac{F^e_{k\rightarrow t}\cdot F^e_t}
{|F^e_{k\rightarrow t}||F^e_t|}\bigg),
\label{Equation 3}
\end{equation}
where $F^e$ is the embedding vector of $F$ outputted by the shallow fully convolutional network. Note that the dot productions in Eq.~(\ref{Equation 2}) and Eq.~(\ref{Equation 3}) is element-wise multiplication and all the obtained weights are normalized so that $\textstyle\sum_{k\in [t-1,t-n]}\omega_{{k}\rightarrow t}=1$.
	
Compared to the existing architectures like TrackR-CNN~\cite{voigtlaender2019mots} which naively uses heavy 3D convolutions for feature extraction and fusion, our method models the spatial movements of objects in successive video frames in a more reasonable fashion to enhance the extracted features, could improve the accuracy for upper-stream tasks. With the help of a lightweight optical flow network~(i.e., FlowNet~\cite{dosovitskiy2015flownet}) to predict the object movements, our method for feature extraction has much fewer parameters than 3D convolutions. Therefore, our method keeps a faster runtime than TrackR-CNN and its variants. In addition, by tuning the temporal range $n$, we can actively control the tradeoff between inferring speed and accuracy.
\vspace{-0.6cm}
\subsection{Associative Connections Across Tasks}
After the above-mentioned flow-guided feature fusion, the obtained feature representation could be significantly enhanced. These enhanced feature maps are then fed into the individual upper-stream task heads. These tasks include object detection, instance segmentation, and object tracking. Although these tasks are intrinsically related, existing methods~(e.g., TrackR-CNN~\cite{voigtlaender2019mots} and CAMOT~\cite{ovsep2018track}) simply add different learning objectives together. This paper explores the relations among these tasks for improving multiple object tracking and segmentation accuracy. Under this motivation, we propose collaborative multi-task learning by adding associative connections among different tasks to facilitate information flow through different heads. Different from previous methods where different task heads are independent, the three heads in our method jointly worked together. To be concrete, as shown in Fig.~\ref{Frame}, there are three associative connections across the detection, segmentation, and tracking heads. The first one is the associative connection between the output of bounding box regression in the detection head and the input of the mask prediction head. The second one is a connection to link the outputs of detection and mask heads. The third one is the associative connection between the combined outputs of detection and mask heads and the input of the tracking head. 
With this implementation, we achieve much lower computational complexity per instance than the mask head in TrackR-CNN, where the ROI-based operations are repeatedly performed for final dense predictions. For comparison, the associative connections facilitate information sharing across different task heads to keep a low computational cost. Meanwhile, owing to the interleaved information flow among all three heads, we can jointly optimize their objectives as a whole, while previous works optimize each task's objective independently since all the task heads are independent in their methods.
	
\subsubsection{Connection from Detection to Segmentation}\label{sec:connect_1}
Due to the good performance of Mask R-CNN~\cite{he2017mask} for instance segmentation, we adopt its head architecture for object detection and segmentation. The detection and instance segmentation are based on region proposals, and our network training procedure contains two stages Mask R-CNN and Faster R-CNN~\cite{ren2015faster}. In the first stage, a Region Proposal Network (RPN) takes the extracted features from the backbone and outputs a set of region proposals for object detection. In the second stage, for the detection head, an ROI pooling layer is used to generate region features for each proposal, which are then used to predict the object class and the related bounding box by the classification $\textbf{N}_{cl}$ and regression $\textbf{N}_{bb}$ heads respectively. Then, for the mask head, an ROI pooling layer is used to generate object-aware features based on the bounding boxes outputted from the detection head. These object-aware features are used by the following layers to conduct instance segmentation for producing object masks. 
	
Adding a connection from the output of the detection head to the ROI pooling layer of the mask head~(i.e., the red arrow between the bounding box and mask heads in Fig.~\ref{Frame}) can explore more accurate features focused on objects compared to using the region proposals that have been widely used in the literature~\cite{voigtlaender2019mots,xu2020pointtrack++}. This is because the output of the detection head is the object bounding boxes that are finer than the region proposals regarding the object locations. Therefore, our mask head can accurately fire on the pixels of instances.
In other words, by adding an associative connection between the output of the detection head and the input of the mask head, we achieve a better behavior for mask generation since the used features could be more focused on the object itself.

The learning objective of our detection head is identical to that of Faster R-CNN,
\begin{equation}
\mathcal{L}_{cls}+\mathcal{L}_{box}=\frac{1}{D}\sum_{d \in D}\mathcal{L'}(p_d, p_d^{*} )+\frac{1}{D}\sum_{d \in D}\mathcal{L''}(t_d, t_d^{*}),
\end{equation}
where $D$ is the total number of detections in a video sequence, $p_d$ is the predicted probability of an object, $p_d^{*}$ is the corresponding ground-truth label,  $t_d$ is the coordinates of the predicted bounding box,  $t_d^{*}$ is the ground-truth for the corresponding bounding box.

Similarly, the instance segmentation head is trained using the following loss function,
\begin{equation}
\mathcal{L}_{mask}=\frac{1}{D}\sum_{d \in D}\mathcal{L'''}(m_d, m_d^{*} ),
\end{equation}
where $m_d$ and $m_d^{*}$ are predicted mask and the ground-truth mask.

Note that the novelty of our method lies in introducing the associative connections to interact among different tasks, not the individual learning objective for each task.

\subsubsection{Connection from Detection and Segmentation to Tracking} \label{sec: dynamicTracking}
	
The tracking head aims to establish correspondences of the same identities across successive frames. Existing works, such as TrackR-CNN, extensively extract ROI features for all region proposals, which are then used to compute identity vectors. Those identity vectors from positive proposals are used to link the same identities across frames based on their similarities. Such a tracking head computes many redundant proposal features and identity vectors. In addition, extracting ROI features from region proposals may not be accurate enough to facilitate good tracking results. To address these issues, we propose to extract ROI features from the detection and instance segmentation results. This not only avoids the redundant feature computation but also provides a more accurate description of the tracked objects. Such an idea inspires the associative connection from the outputs of detection and segmentation heads to the input of the tracking head.
	
Usually, the bounding boxes generated by the detection head are larger than the ground truth boxes, while the bounding boxes computed from the masks produced by the mask head are often tighter than the ground truth ones due to the pixel-level prediction of the mask head. To effectively combine two kinds of bounding boxes for a better ROI feature extraction in the tracking head, we propose an adaptive fusion strategy to obtain a weighted bounding box as,
\begin{equation}
b^i_{wb}=\alpha_1 \cdot b^i+\alpha_2 \cdot b^i_{mk},
\label{Equation 4}
\end{equation}
where $b^i$ is the bounding box outputted by the detection head and $b_{mk}^i$ is the bounding box drawn from the predicted mask for the $i^{th}$ object, $\alpha_1$, $\alpha_2$ are adaptive weights fulfilling $\alpha_1+\alpha_2=1$. \textcolor{black}{The parameters $\alpha_1$ and $\alpha_2$ are hyperparameters but they do not vary for each scenario right now. We will investigate how to adjust them for each video sequence in future work.}
	
In this way, the weighted bounding boxes $b_{wb}$ will be smaller than the bounding boxes $b$ generated by the detection head, while larger than the mask-based bounding boxes $b_{mk}$. If the object is on a large scale, we expect the weighted bounding box to be tighter for tracking. However, if the object is small, the mask-based bounding box may not be accurate enough so we expect the weighted bounding box to approach the detected bounding box to include more information. Under this consideration,  $\alpha_1$ is set proportional to the reciprocal of the scale of the object, to be specific, the area of bounding box ${b}^{i}$. Therefore, we have a small $\alpha_1$ ($\alpha_2$ is large) and ${b}^{i}_{mk}$ has a larger impact on the weighted bounding box ${b}^{i}_{wb}$ when the object is large. On the contrary, ${b}^{i}$ will have a larger impact on ${b}^{i}_{wb}$ when the object is small. 
	
Owing to the good properties of these adaptive bounding boxes, we add an associative connection to link these boxes into the tracking head for ROI feature extraction. In other words, we use $b_{wb}$ to pool feature maps to be more object-aware. Following the idea of~\cite{voigtlaender2019mots,tang2017multiple} that use the embedding vectors to re-identify persons, our tracking head uses one fully connected layer to map ROI features into identity vectors $v$, each of which indicates a unique identified instance. These identity vectors $v$ are used to link all detections across frames. Since our ROI features are extracted from object-aware feature maps and could be more discriminative for tracking multiple objects, our tracking head can predict more distinguishable identity vectors.
	
In the training stage, we minimize the distances of all vectors belonging to the same object while maximizing the distances of vectors belonging to different objects. To this end, the tracking head is trained with the batch hard triplet ranking loss. For each detected object, we sample its hard positive detections and negative detections for network training. Let us denote $D$ as all the detections in a video. At frame, $t$, each detection $d \in D$ is associated with a tracking vector $v_d$ and a ground truth track $ID$ which is used to determine its overlap with the ground truth over frames. Thus, for a video sequence including $T$ frames, the tracking loss is computed by,
	
\begin{equation}
\begin{aligned}
\mathcal{L}_{track} &= \frac{1}{D}\sum_{d \in D}\max\bigg(m + \max_{\substack{n \in D \\ id_{d} \neq id_{n}}} \mathcal{S}(v_d,v_n) \\
&- \min_{\substack{p \in D \\ id_{d}=id_{p}}} \mathcal{S}(v_d,v_p),0\bigg),
\end{aligned}
\label{Equation 6}
\end{equation}
where the subscript $p$ and $n$ indicate the positive and negative detections respectively \textcolor{black}{ and $\mathcal{S}$ represents the similarity between the input vectors. By default, we use cosine similarity.}
\vspace{-0.4cm}
\subsection{Inference and Training}
In this section, we first elaborate on the inference algorithm, then analyze the runtime complexity, and finally discuss the training objective of the proposed method.
\subsubsection{Inference algorithm}
We summarize the inference of our proposed method in Algorithm 1. The algorithm of our proposed method can be divided into two major stages. In the first stage, given a video frame $I_t$, the feature extraction network $\textbf{N}_{fm}$ takes it as input and produces a set of intermediate feature maps $F_t$~(line 2 in Algorithm 1). Based on the light-weight optical flow network $\textbf{N}_{fl}$, we warp the temporal feature maps in previous frames to the current frame~(line 3 to 4) according to Eq.~(\ref{Equation 1}). In the meantime, we calculate the similarities of the warped feature maps to that of the current frame. These similarities are used as adaptive weights~(line 5) to perform flow-guided feature fusion over the temporally aligned features, to enhance the feature representation for the current frame. In this way, the enhanced feature maps $F'_t$~(lines 6 to 7) can be obtained for different head tasks. In the second stage, the enhanced feature maps $F'_t$ are fed into the classification and bounding box heads to predict object class $\{c_t\}$ and bounding box $\{b_{t}\}$ respectively~(line 8). Using the predicted bounding box from $\{b_{t}\}$, we further obtain the object-aware feature maps $F'_{(bb)t}$ which are more accurate than the proposal-based features that are widely used in Mask RCNN~\cite{he2017mask} and its variants~\cite{voigtlaender2019mots,xu2020pointtrack++}. Next, the feature maps $F'_{(bb)t}$ are fed into the mask head $\textbf{N}_{mk}$~(line 11). Since the mask head $\textbf{N}_{mk}$ takes the object-aware feature maps as inputs, it tends to generate masks~(denoted as $\{m_t\}$) with higher quality compared to those masks predicted from proposal based features. We draw a close rectangular over each instance in $\{m_t\}$ to obtain a mask-based bounding box $\{b_{(mk)t}\}$.
Considering the predictions from both $\{b_{t}\}$ and $\{b_{(mk)t}\}$ together, we compute adaptive weights to get a weighted bounding box $\{b_{(wk)t}\}$ for each detected object~(line 12) which could be more accurate than the original bounding boxes $\{b_t\}$ produced by the bounding box head. Intuitively, more accurate bounding boxes can benefit following object tracking heads as they only focus on the major component of objects, reducing the influence of background.  The final video recognition result $R_{t}$ consists of a set of object classes $\{c_t\}$ , bounding-box locations $\{b_t\}$, object masks $\{m_t\}$, and tracked identity labels $\{t_t\}$.
	
\begin{table}
    \centering
    \begin{tabular}{p{0.95\columnwidth}}
        \toprule
        \textbf{Algorithm 1} Online inference of the proposed method per frame\\
        \midrule
        1: \textbf{input}: frame $ \{I_t\} $ \hfill $\triangleleft$ Video frame at time $t$\\
        2: $F_t = \textbf{N}_{fm}\big(I_t\big)$ \hfill $\triangleleft$ Produce feature maps\\
        3: \textbf{for} $j = t-n$ \textbf{to} ${t-1}$  \textbf{do} \hfill $\triangleleft$ Use temporal features\\
        4:\hspace{0.4cm} $F_{j\rightarrow t} = \mathcal{WP}\Big(F_j, \textbf{N}_{fl}\big(I_j, I_t\big)\Big)$ \hfill$\triangleleft$ Feature warping to time $t$\\
        5: \hspace{0.4cm}    $\omega_{j\rightarrow t}= exp\Big(
        \frac{\textstyle F^e_{j\rightarrow t}\cdot F^e_t}
        {\textstyle |F^e_{j\rightarrow t}||F^e_t|}\Big)$ \hfill$\triangleleft$ Calculate adaptive weights\\
        6: \textbf{end for}\\
        7: $ F'_{t} =\textstyle\sum^{t-1}_{j=t-n}\big(\omega_{j\rightarrow t}\cdot F_{j\rightarrow t}\big)+F_t$ \hfill$\triangleleft$ Flow-guided feature fusion\\
        8: $\substack{\{c_t\}= \textbf{N}_{cl}\big(F'_{t}\big)\\ \{b_{t}\}= \textbf{N}_{bb}\big(F'_{t}\big)}$ \hfill$\triangleleft$ Object detection results\\
        9: $F'_{(bb)t}=\{b_{t}\} \rightarrow F'_{t}$ \hfill$\triangleleft$  bounding-box features maps\\
        10: $\{m_{t}\}= \textbf{N}_{mk}\big(F'_{(bb)t}\big)$ \hfill$\triangleleft$ Produce mask results\\
        11: $b_{(mk)t} \leftarrow \{m_{t}\} $ \hfill$\triangleleft$  Mask-based bounding box\\
        12: $\{b_{(wb)t}\}=\sum\limits_{i \in \mathbb{R}^{n}} \alpha_1 \cdot b^i_{t}+\alpha_2 \cdot b^i_{(mk)t}$ \hfill$\triangleleft$ Weighed bounding box\\
        13: $F'_{(tb)t}=\{b_{(wb)t}\} \rightarrow F'_{t}$ \hfill$\triangleleft$  Object-aware features maps\\
        14: $\{t_{t}\}= \textbf{N}_{tr}\big(F'_{(tb)t}\big)$ \hfill$\triangleleft$ Produce tracking results\\
        13: $R_{t}=\Big(\{b_t\},\{c_t\},\{m_t\},\{t_{t}\}\Big)$
        \hfill$\triangleleft$ 
        Final recognition at $t$\\
        15: \textbf{output}: Video recognition result $R_t$.  \\
        \bottomrule
    \end{tabular}
    \vspace{-0.6cm}
\end{table}
\subsubsection{Time complexity}\label{sec:time}
Based on Algorithm 1, we give an analysis of the time complexity of the proposed method. Besides the backbone feature extraction network $\textbf{N}_{fm}$, there are five modules: (1) The optical flow network $\textbf{N}_{fl}$ which includes the bilinear warping and feature embedding functions; (2) The classification head $\textbf{N}_{cl}$; (3) The bounding-box regression head $\textbf{N}_{bb}$; (4) The mask head $\textbf{N}_{mk}$; (5) The tracking head $\textbf{N}_{tr}$. Since our associative connections are parameters-free and only work as information flow which is fast, we exclude it in our analysis. Given a temporal range of $n$ in flow-guided feature fusion, the optical flow network $\textbf{N}_{fl}$ loops $n$ times per frame for feature warping. Accordingly, the runtime complexity for the proposed method is,
\begin{equation}
\begin{aligned}
\small
\mathcal{O}_{ours} &=\mathcal{O}\left(\textbf{N}_{fm}\right)+n\cdot\mathcal{O}\left(\textbf{N}_{fl}\right)+\mathcal{O}\left(\textbf{N}_{cl}\right)+\mathcal{O}\left(\textbf{N}_{bb}\right) \\
&+\mathcal{O}\left(\textbf{N}_{mk}\right)+\mathcal{O}\left(\textbf{N}_{tr}\right)
\label{Equation 8}
\end{aligned}
\end{equation}
	
We compare the time complexity of our method to its predecessor, TrackR-CNN~\cite{voigtlaender2019mots} as both of them have similar architecture and TrackR-CNN is the state-of-the-art multi-object tracking and segmentation. With the same symbols, TrackR-CNN has the following runtime complexity,
\begin{equation}
\begin{aligned}
\small
\mathcal{O}_{tr} &=\mathcal{O}\left(\textbf{N}_{fm}\right)+ m\cdot\mathcal{O}\left(\textbf{N}_{3D}\right)+\mathcal{O}\left(\textbf{N}_{cl}\right) \\
&+\mathcal{O}\left(\textbf{N}_{bb}\right)+\mathcal{O}\left(\textbf{N}_{mk}\right)+\mathcal{O}\left(\textbf{N}_{tr}\right),
\label{Equation 9}
\end{aligned}
\end{equation}
where $\mathcal{O}\big(\textbf{N}_{3D}\big)$ is the 3D convolutions and $m$ is the feature fusion length used in TrackR-CNN for per frame feature extraction. 
	
Typically, the backbone feature extraction network has a heavier architecture than the task heads because it contains many more layers for feature abstraction. Therefore, it is reasonable to assume $\mathcal{O}\big(\textbf{N}_{cl}\big) \ll \mathcal{O}\big(\textbf{N}_{fm}\big)$, $\mathcal{O}\big(\textbf{N}_{bb}\big) \ll \mathcal{O}\big(\textbf{N}_{fm}\big)$, $\mathcal{O}\big(\textbf{N}_{mk}\big) \ll \mathcal{O}\big(\textbf{N}_{fm}\big)$, $\mathcal{O}\big(\textbf{N}_{tr}\big) \ll \mathcal{O}\big(\textbf{N}_{fm}\big)$,  and $\mathcal{O}\big(\textbf{N}_{mk}\big) \ll \mathcal{O}\big(\textbf{N}_{fm}\big)$. Given these, the ratio of runtime complexity of the proposed method to TrackR-CNN~\cite{voigtlaender2019mots} can be computed as:
\begin{equation}
C=\frac{\mathcal{O}_{ours}}
{\mathcal{O}_{tr}} \approx \frac{\mathcal{O}\big(\textbf{N}_{fm}\big)+n\cdot\mathcal{O}\big(\textbf{N}_{fl}\big)}
{\mathcal{O}\big(\textbf{N}_{fm}\big)+m\cdot\mathcal{O}\big(\textbf{N}_{3D}\big)}.
\label{Equation 10}
\end{equation}
	
Compared to $\textbf{N}_{3D}$ used in TrackR-CNN, we use $\textbf{N}_{fl}$ which is more efficient while still being effective for object movement modeling. Assuming $n=m$, $C$ in Eq.~(\ref{Equation 10}) is less than 1. Therefore, our method is faster than TrackR-CNN. Experimental results in Table~\ref{tab: KITTI MOTS} also demonstrate that our method is two times faster than TrackR-CNN.
	
\subsubsection{Training objective}
Since our method simultaneously considers object detection, segmentation, and tracking, its training objective contains multiple losses accordingly,
\begin{equation}
\mathcal{L}_{total}=\mathcal{L}_{cls}+\mathcal{L}_{box}+\mathcal{L}_{mask}+\mathcal{L}_{track},
\label{Equation 11}
\end{equation}
where $\mathcal{L}_{cls}$, $\mathcal{L}_{box}$, $\mathcal{L}_{mask}$, and $\mathcal{L}_{track}$ denotes the loss for classification, bounding box regression, mask segmentation, and object tracking respectively. Following the implementation from Mask R-CNN\cite{he2017mask}, we modify the classification loss $\mathcal{L}_{cls}$ and bounding-box loss $\mathcal{L}_{box}$ based those from Faster R-CNN\cite{ren2015faster}. \textcolor{black}{In detail, we add extra losses before and after the associative connections across tasks for full supervision. Therefore, both the predictions before and after the associative connections are optimized.} As we use a set of $m \times m$ for mask generation, the mask head has a $cm^2$-dimensional output for each ROI, which encodes $c$ binary masks. To achieve this goal, we use a per-pixel sigmoid, and $\mathcal{L}_{mask}$ is defined as the average binary cross-entropy loss. For an ROI associated with ground-truth class $k$, we only define $\mathcal{L}_{mask}$ on the $k^{th}$ mask while ignoring other mask outputs that do not contribute to the loss. The details of $\mathcal{L}_{track}$ are discussed in Section~\ref{sec: dynamicTracking}.
	
As the entire framework of the proposed method is multiple-stage, we rely on predictions from the classification head for the label of object detection, instance segmentation (mask), and tracking. Using the classification result for each ROI, we allow each task head to generate its results. Namely, the bounding-box head, the mask head, and the tracking head only need to focus on their specific task based on ROI without competition among classes. With the associative connections, critical ROI information can forward pass from the bounding box head to the mask head and eventually to the tracking head respectively. In the same vein, the ground truth for instance segmentation and object tracking can be backpropagated to each task head in the training procedure. 
\vspace{-0.4cm}
\section{Experiments}
This section first describes the datasets and evaluation metrics that we use to assess our method. The implementation details with training parameter settings are then elaborated. Next, We supply ablation studies to investigate the improvements of our method from a strong baseline and its variants. We also compare our method with the state-of-the-art methods on two benchmarks. Finally, we conclude with the accuracy and runtime evaluation.

 \begin{table*}[ht]
    \caption{Comparison with the state-of-the-art methods on the KITTI MOTS dataset.}
    \vspace{-0.3cm}
    \label{tab: KITTI MOTS}
\centering
\begin{tabular}{c|c|c|c|c|c|c|c}
    \toprule    
    \multirow{2}{*}{Method} & \multirow{2}{*}{Detect + Segment} & Speed &   \multirow{2}{*}{\textcolor{black}{FPS}} & \multicolumn{4}{c}{Cars / Pedestrians}  \\
    \cline{5-8}
    & & (s) & & sMOTSA$\uparrow$ & MOTSA$\uparrow$ & MOTSP$\uparrow$ & IDS$\downarrow$\\
    \midrule
    CAMOT \cite{ovsep2018track} & TRCNN & 0.76 & 1.32 & 67.4 / 39.5 & 78.6 / 57.6 & 86.5 / 73.1& 220 / 131 \\
    CIWT\cite{osep2017combined} & TRCNN & 0.28 & 3.57 & 68.1 / 42.9  & 79.4 / 61.0 & 86.7 / 75.7 & 106 / 42 \\
    ReMOTS\cite{yang2020remots} & TRCNN + BB2SegNet & 3.33 / - & 0.30 & 70.4 / -& 84.4 /- & - / - & 231 / -\\
    TrackR-CNN\cite{voigtlaender2019mots} & TRCNN & 0.50 & 2.00 & 76.2 / 46.8 & 87.8 / 65.1 & 87.2 / 75.7 & 93 / 78 \\
    BePix \cite{BeyondPixels_ICRA2018} & RRC\cite{ren2017accurate} + TRCNN & 0.36 & 2.77 & \textbf{76.9} / - & \textbf{89.7} / - & 86.5 / - & 88 / - \\
    \textcolor{black}{Stem-Seg} \cite{athar2020stem} & TRCNN & 0.32 & 3.12 & 72.7 / \textbf{50.4} & 83.8 / \textbf{66.1} & 87.2 / \textbf{77.7} & 76 / \textbf{14}  \\
    \midrule
    Ours & TRCNN & \textbf{0.27} & \textbf{3.70} & 76.7 / 47.9 & 88.2 / {65.3} & \textbf{88.5} /76.1 & \textbf{62} / 32\\
    \bottomrule
\end{tabular}
\vspace{-0.6cm}
\end{table*}
\vspace{-0.4cm}
\subsection{Datasets and Evaluation Metrics}
KITTI MOTS~\cite{voigtlaender2019mots} and MOTS Challenge~\cite{milan2016mot16} are used to evaluate the effectiveness of the proposed method. KITTI MOTS has 21 video sequences of 8,008 frames, while MOTSChallenge has four video sequences of 2,862 frames. For KITTI MOTS, it includes 11,420 pedestrian and 26,899 car instances. There are 26,894 pedestrian instances in the MOTS Challenge.
	
Following TrackR-CNN~\cite{voigtlaender2019mots}, we use \textbf{MOTSA} to measure the accuracy for multi-object tracking and segmentation, \textbf{MOTSP} to measure the precision of mask-based multi-object tracking and segmentation results, and \textbf{sMOTSA} to evaluate the soft multi-object tracking and segmentation accuracy. They are defined as follows,
\begin{equation}
\begin{aligned}
MOTSA&= \frac{|TP|-|FP|-|IDS|}
{|M|},\\
MOTSP&= \frac{|\widetilde{TP}|}
{|TP|},\\
sMOTSA&= \frac{\widetilde{TP}-|FP|-|IDS|}
{|M|},
\label{Equation 12}
\end{aligned}
\end{equation}
where $TP$, $\widetilde{TP}$, $FP$, and $IDS$ are true positive, soft true positive, false positive, and tracking ID switch score respectively. We refer readers to \cite{voigtlaender2019mots} for more details about each notation. These metrics collectively measure the performance of a multi-object tracking and segmentation system by considering three tasks~(i.e., object detection, instance segmentation, and multi-object tracking) together.
\vspace{-0.5cm}
\subsection{Implementation Details}
The proposed method is implemented on a workstation with one NVIDIA RTX GPU. To have a fair comparison with existing methods, we follow the same experimental setup as in TrackR-CNN~\cite{voigtlaender2019mots}. To be specific, the backbone feature extraction network ResNet-101~\cite{he2016deep} is pretrained on COCO~\cite{lin2014microsoft} and Mapillary~\cite{MVD2017} datasets. The optical flow network FlowNet is pretrained on the Flying Chair dataset~\cite{dosovitskiy2015flownet}. During the training process, the weights of ResNet-101 and FlowNet are fixed, and the other weights related to different task heads~(i.e.,  ${\textbf{N}}_{bb}$, ${\textbf{N}}_{cl}$, ${\textbf{N}}_{mk}$ and ${\textbf{N}}_{tr}$) are updated by learning on the target dataset, $i.e.$ KITTI MOTS or MOTS Challenge. We train our model for 40 epochs with a learning rate of $5 \times {10}^{-7}$ using the Adam~\cite{kingma2014adam} optimizer and mini-batch size of eight. The temporal range $n$ in Eq.~(\ref{Equation 2}) is set to eight, i.e., eight adjacent frames are used for the flow-guided feature extraction. 
	
For the KITTI MOTS benchmark, there are 21 videos in total and we use 12 of them for training and the remaining for testing, following the practice in TrackR-CNN. We randomly keep some training data for validation and choose the best model on the validation set for testing. For the MOTS Challenge benchmark, since there are only four video sequences in total, we use cross-validation to test the performance of different methods. To be specific, we leave one video sequence for evaluation and train the model on the three others on the MOTS Challenge. This process is repeated four times and the average result is reported. 
\vspace{-0.4cm}
\subsection{Comparison with the State-of-the-art Methods}
\noindent\textbf{Results on KITTI MOTS.} The results compared with the state-of-the-art methods on KITTI MOTS dataset are shown in Table~\ref{tab: KITTI MOTS}. For a fair comparison, we list the leading methods with detection and segmentation architectures based on TrackR-CNN, which is the same as ours. The best result for each metric is highlighted in the table. The results of our method in Table~\ref{tab: KITTI MOTS} is quite encouraging. For car recognition, our method achieves the best result on MOTSP and IDS. Especially, our method improves IDS over other methods by a significantly large margin, which leads to the second-best method by 26. sMOTSA and MOTSA of our method are on par with the best model, a.k.a., BePix~\cite{BeyondPixels_ICRA2018}. Regarding speed, our method is faster than BePix. The fast inference speed of our method is due to its flow-guided feature extraction in which a lightweight flow network is used. Finally, for pedestrian recognition, our method is better than all the compared methods on these metrics.

\begin{figure*}[!ht]
    \centering
    \includegraphics[width=1\textwidth]{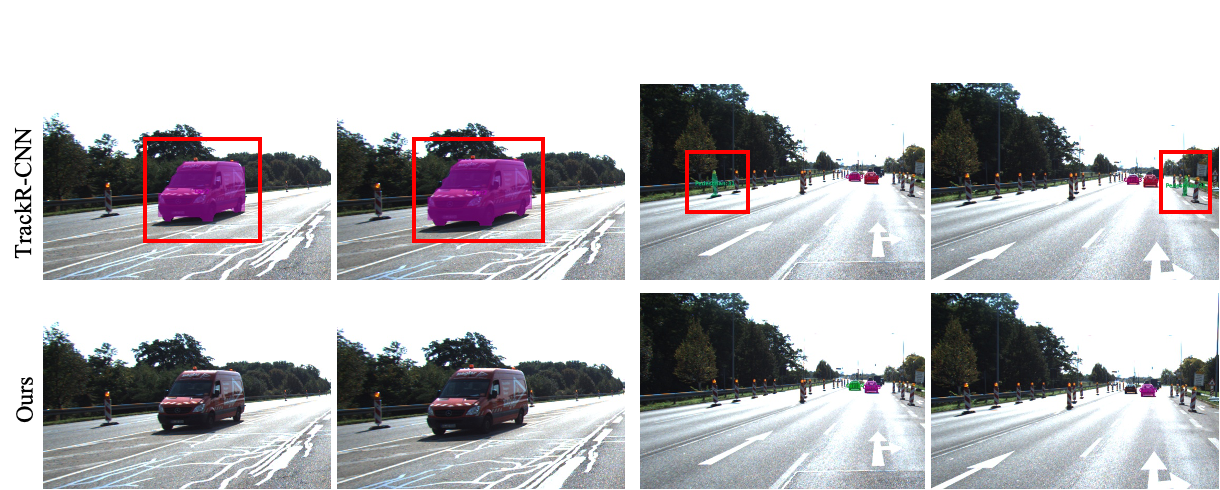}
    \vspace{-0.4cm}
    \caption{Qualitative comparison between TrackR-CNN and our method regarding false prediction. The false predictions are marked by red bounding boxes. Minivan is not an object in the KITTI MOTS dataset. In the first two rows, TrackR-CNN has a false prediction on the minivan as a car class. In the same vein, TrackR-CNN has a false recognition of traffic signs and predicts them as "pedestrian". For both scenarios, our method works correctly without false predictions. The images are cropped for better visualization.}
    \label{mis}
\end{figure*}

\begin{figure*}[!ht]
    \centering
    \includegraphics[width=\textwidth]{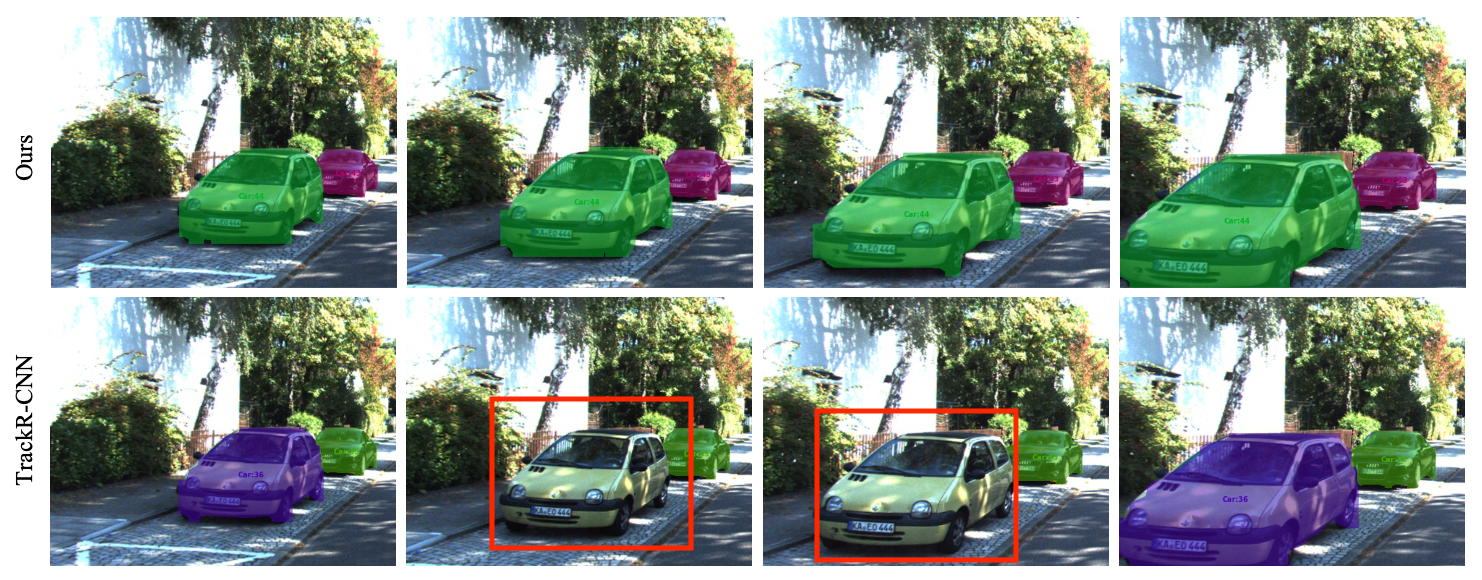}
    \vspace{-0.4cm}
    \caption{Comparison of TrackR-CNN and our results on missed recognition. The missed recognition is marked in red bounding boxes. Our method can constantly perform recognition on video frames while TrackR-CNN encounters recognition lost for the middle two frames. The demonstrated examples are cropped for better visualization.}
    \label{lost}
\end{figure*}

To qualitatively demonstrate the improvements of our method, we visualize our results and compare them with TrackR-CNN. As shown in Fig.~\ref{mis}, our method is less prone to false predictions. In the left two columns of Fig.~\ref{mis}, TrackR-CNN in the top row produces a "car" prediction on the red minivan while our method in the bottom row does not. A minivan is not a labeled object in the KITTI MOTS dataset. For the right two columns of Fig.~\ref{mis}, TrackR-CNN produces a false detection on traffic signs, predicting them as "pedestrian". In these examples, our method performs accurate predictions for both scenarios. What is more, TrackR-CNN frequently encounters missing detection. For comparison, our method can constantly detect objects through video frames as demonstrated in Fig.~\ref{lost}. The improvement of our method over TrackR-CNN mainly stems from the reliable object-aware features, which are brought from our associative connections. With the help of these connections, our method produces better feature representations than TrackR-CNN before the task heads produce individual predictions, thus achieving better results.

We also observe that our method has improved performance on tracking and mask quality. We argue that the proposed associative connections benefit the mask and tracking predictions as for the two tasks, we use refined ROI features for the final prediction. In contrast, TrackR-CNN produces predictions using coarse features based on proposals.

\noindent\textbf{Results on MOTS Challenge.} Table~\ref{tab:MOTSChallenge} reports the results on the MOTS Challenge dataset. Our method outperforms all the other methods for all metrics. The improvements over the second best method~(TrackR-CNN), are 0.5 on sMOTSA, 0.2 on MOTSA, and 0.7 on MOTSP respectively. Besides TrackR-CNN, our method achieves significantly better results than other methods even if they use a domain-finetuned Mask R-CNN.
	
\begin{table}[!bt]
\vspace{-0.4cm}
    \caption{Comparison with the state-of-the-art methods on MOTS Challenge dataset. \textcolor{black}{+MG denotes mask generation with a domain finetuned Mask R-CNN.}}
    \vspace{-0.3cm}
    \label{tab:MOTSChallenge}
    \centering
    \begin{tabular}{c|c|c|c}
        \toprule
        Method & sMOTSA$\uparrow$ & MOTSA$\uparrow$ & MOTSP$\uparrow$\\
        \midrule
        MOTDT\cite{Chen_2018} + MG & 47.8 & 61.1 & 80.0 \\
        MHT-DAM\cite{kim_ICCV2015_MHTR} + MG & 48.0 & 62.7 & 79.8 \\
        jCC\cite{Keuper2020MotionS} + MG & 48.3 & 63.0 & 79.9 \\
        FWT\cite{henschel2018fusion} + MG & 48.3 & 64.0 & 79.7 \\
        TrackR-CNN\cite{voigtlaender2019mots} & 52.7 & 66.9 & 80.2\\
        \midrule
        Ours & \textbf{53.2} & \textbf{67.1} & \textbf{80.9} \\
        \bottomrule
    \end{tabular}
    \vspace{-0.4cm}
\end{table}

\begin{figure*}[!ht]
    \centering
    \includegraphics[width=\textwidth]{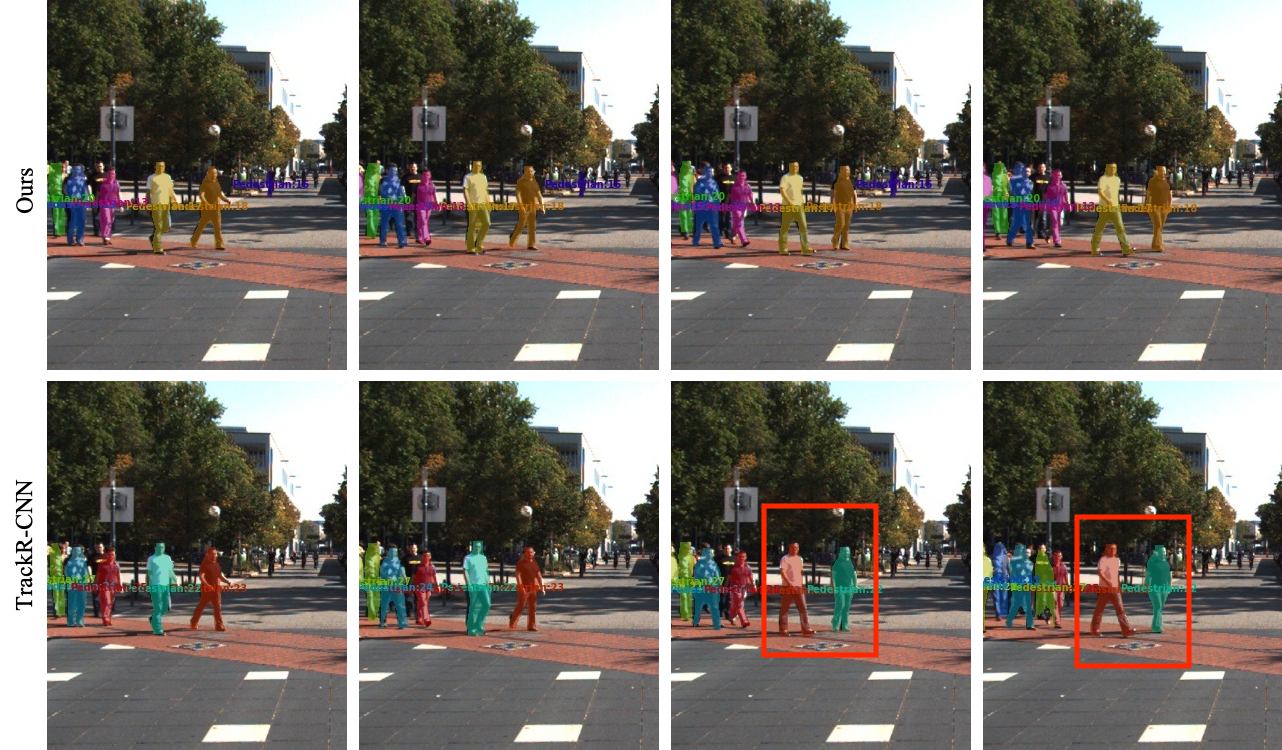}
    \vspace{-0.6cm}
    \caption{Comparison of TrackR-CNN and our results on ID switch. The incorrect cases for tracking are marked in red bounding boxes. Our method can constantly perform accurate tracking of the same instances on video frames while TrackR-CNN frequently produces incorrect predictions. The demonstrated examples are cropped for better visualization.}
    \label{ids}
    \vspace{-0.4cm}
\end{figure*}
\vspace{-0.4cm}
\subsection{Ablation Study}
To demonstrate the effectiveness of our design, we conduct ablation studies from TrackR-CNN since our method is proposed for the same purpose as TrackR-CNN and shares a similar architecture to it. Specifically, we first replace the computationally expensive conv3d in TrackR-CNN with more efficient flow-guided feature extraction proposed in Section~\ref{sec:feat}, which leads to Method A. Then, we add an associative connection from the bounding box head to the mask head as described in Section~\ref{sec:connect_1} and obtain Method B. By further adding an associative connection from the bounding-box head to track, this leads to Method C in which bounding box features are used for both segmentation and tracking, rather than the proposal based features. A variant of Method C is to connect the mask-based bounding box to the tracking head, which we call Method D. We also test two kinds of strategies to fuse bounding boxes and mask-based bounding boxes, which are then input to the tracking head. Method E uses fixed weight $\alpha=0.5$ to fuse these two kinds of bounding boxes, while Method F is the proposed one that fuses them with adaptive weights.
	
Table~\ref{tab:ablation} demonstrates the results of each method on the KITTI MOTS dataset. For reference, we also include the results of TrackR-CNN. By comparing A to TrackR-CNN, we can find that using flow-guided feature extraction leads to slightly worse accuracy, however, it leads to faster runtime as shown in Table~\ref{tab: KITTI MOTS}. Based on A, B obtains better results, proving our associative connection that using predicted bounding boxes is better than region proposals in pooling features for the mask head. C and D further improve B, demonstrating the effectiveness of introducing an associative connection to the tracking head. If these two kinds of bounding boxes together, better results are obtained. More importantly, the proposed adaptive box fusion strategy~(Method F) achieves significant improvement over the hard fusion~(Method E), especially for the tracking consistency. These results show that the proposed method effectively explores the properties of these two kinds for multi-object tracking and segmentation. 
\begin{table*}[ht]
\vspace{-0.4cm}
	\caption{Ablation study on the KITTI MOTS dataset. Please see the texts for details about methods A to E, and F is the proposed method.}
	\vspace{-0.3cm}
 \label{tab:ablation}
	\centering
	\begin{tabular}{c|c|c|c|c|c|c|c|c}
		\toprule
		\multicolumn{2}{c|}{Method} & TrackR-CNN & A & B & C & D & E & F \\
		\midrule
		\multirow{4}{*}{Cars} & sMOTSA & 76.2 & 75.6 & 76.0 & 76.1 & 76.5 & 76.3 & 76.5  \\
		& MOTSA & 87.8 & 87.0 & 87.1 & 87.5 & 88.1 & 87.9 & 88.2 \\
		& MOTSP & 87.2 & 86.9 & 87.1 & 87.4 & 88.4 & 88.3 & 88.5 \\
		& IDS & 93 & 95 & 87 & 76 & 74 & 73 & 62 \\
		\midrule
		\multirow{4}{*}{Pedestrians} & sMOTSA & 46.8 & 45.2 & 45.5 & 46.1 & 46.1 & 46.2 & 47.9 \\
		& MOTSA & 65.1 & 63.5  & 63.8 & 64.3 & 64.3 & 64.4 & 65.3 \\
		& MOTSP & 75.7 & 75.0 & 75.1 & 75.8 & 75.9 & 76.0 & 76.1 \\
		& IDS & 78 & 76 & 43 & 39 & 38 & 36 & 32 \\
		\bottomrule
	\end{tabular}
 \vspace{-0.4cm}
	\end{table*}
\vspace{-0.4cm}	
\subsection{Fusion Length and Accuracy}
We use eight frames as the default temporal range for feature fusion based on the object movements across frames. To evaluate the impact of this fusion length on the accuracy, we change the number of input frames for our method and test our method on the KITTI MOTS dataset. Table~\ref{tab:ablationSpeed} reports the results concerning the number of input frames. For both cars and pedestrians, the recognition accuracy increases when the temporal range increases from 2 to 8. After that, the performance starts to degrade. Such results are reasonable, because only a few frames may not accumulate enough information to enhance the feature of the current frame while too many frames are also damageable as the long-term movements estimated by the optical flow network are unstable. In addition, it is also worth pointing out that a larger temporal range requires more time to process feature extraction, thus will result in a lower speed. Therefore, we need to pay more attention to the temporal range in practice to achieve a desirable recognition accuracy as well as the runtime speed. 
	
\begin{table}[htbp]
\vspace{-0.4cm}
    \caption{Influence of the temporal range used in the flow-guided feature fusion. The results are on the KITTI MOTS dataset.}
    \vspace{-0.3cm}\label{tab:ablationSpeed}
    \centering
    \begin{tabular}{c|c|c|c|c|c|c}
        \toprule
        \multicolumn{2}{c|}{temporal range $n$} & 2 & 4 & 8 & 12 & 16 \\
        \midrule
        \multirow{4}{*}{Cars} & sMOTSA$\uparrow$ & 75.2 & 75.7 & 76.7 & 75.4 & 75.3\\
        & MOTSA$\uparrow$ & 86.5 & 87.6 & 88.2 & 86.8 & 86.7 \\
        & MOTSP$\uparrow$ & 87.3 & 87.4 & 88.5 & 87.8 & 87.2\\
        & IDS$\downarrow$ & 87 & 75 & 62 & 73 & 78\\
        \midrule
        \multirow{4}{*}{Pedestrians} & sMOTSA$\uparrow$ & 46.2 & 46.9 & 47.9 & 46.5 & 46.3 \\
        & MOTSA$\uparrow$ & 63.5 & 64.3 & 65.3 & 63.7 & 63.6\\
        & MOTSP$\uparrow$ & 75.4 & 75.7 & 76.1 & 75.5 & 75.4\\
        & IDS$\downarrow$ & 39 & 36 & 32 & 38 & 35\\
        \bottomrule
    \end{tabular}
    \vspace{-0.6cm}
\end{table}	
	
\vspace{-0.4cm}
\section{Conclusion}
In this work, we propose a  novel algorithm capable of associatively detecting,  segmenting, and tracking multiple objects for video analysis. By adding associative connections across detection, segmentation, and tracking heads in an end-to-end learnable CNN, our method enables information propagation through different tasks, which could benefit all the considered tasks. Therefore, our method achieves state-of-the-art performance on various metrics regarding object detection, instance segmentation, and multi-object tracking, according to our evaluation of two benchmarks. We also conducted extensive ablation studies to demonstrate the effectiveness of each associative connection.
	


\bibliographystyle{IEEEtran}
\bibliography{egbib}

\end{document}